# Limiting Network Size within Finite Bounds for Optimization

*Linu Pinto*[a*], *Dr.Sasi Gopalan*[b]

[a] *Research Scholar, Department of Mathematics, Cochin University of Science and Technology, Kochi-682022, Kerala, India*
[b] *Professor in Mathematics, School of Engineering, Cochin University of Sience and Technology, Kochi-682022, Kerala, India*

*Keywords:*

Feed Forward Neural Network
Back propagation Algorithm
VC Dimension
Sigmoid Networks
Dissimilarity Measure

A B S T R A C T

Largest theoretical contribution to Neural Networks comes from VC Dimension which characterizes the sample complexity of classification model in a probabilistic view and are widely used to study the generalization error. So far in the literature the VC Dimension has only been used to approximate the generalization error bounds on different NN architectures. VC Dimension has not yet been implicitly or explicitly stated to fix the network size which is important as the wrong configuration could lead to high computation effort in training and leads to overfitting. So there is a need to bound these units so that task can be computed with only sufficient number of parameters. For binary classification tasks shallow networks are used as they have universal approximation property and it is enough to size the hidden layer width for such networks. The paper brings out a theoretical justification on required attribute size and its corresponding hidden layer dimension for a given sample set that gives an optimal binary classification results with minimum training complexity in a single layered feed forward network framework. The paper also establishes proof on the existence of bounds on the width of the hidden layer and its range subjected to certain conditions. Findings in this paper are experimentally analyzed on three different datasets using Mathlab 2018 (b) software.

## 1. INTRODUCTION

Deep neural Network have successfully made its application in many pattern recognition problems [1], [2] and hence its theoretical properties especially the expressive power in terms of network complexity are currently an active area of research. Even though studies [3] shows that deep architectures are more compact in representing complex functions compared to shallow networks. There are empirical results [4] that shows the ability of shallow networks in identifying some complex functions as deep networks with same number of parameters. But the theoretical analysis complimenting these results are still an open question that yet need to be studied. Although topological justification [5],[6],[7],[8] has been done on the theoretical analysis but they demand an exponential increase in the number of parameters.

It has long been known from the studies [9],[10] that shallow networks is suficient to compute and approximate any complex function on a compact domain upto a desired level of accuracy and are well known to have universal approximation property. But theorems supporting their claim do not give an estimate on the network size and assume infinite number of network units. So far studies were done on the reduction of this network complexity by use of many regularization techniques like weight decay [11]. The approximation capabilities and generalization error that controlled by measures from statistical learning theory provides a good theoretical explanation on the network complexity in terms of sample size and its distribution. This technique was quantified by a concept VC Dimension by Vapnik and Chervonenkis in 1971 [12]. This concept was widely used to compute generalization error bounds on the architecture on the deep and shallow networks and are not explicitly or implicitly applied to configure the network complexity.

In this paper we investigate the adaptability of VC Dimension to structure shallow networks architecture by limiting the network size to a finite value thereby balancing the dimensions of the sample space.

### 1.1 Literature Review

Arguments on the suitable depth and width of a network is a widely discussed open problem [12], [14], [15], [16], [17], [18], [19], [20], [21]. Many of these recent papers demonstrate that increasing the depth of a network leads to the reduction in the dimension of each layer. A Feed forward network with a single hidden layer could approximate any continuous function or a measurable function on a compact set with a continuous sigmoidal activation function [22], [23], [24].

Later many results on Feed Forward Neural Network were proved with various methods in [25], [26], [27], [28], [29], [30], [31], [32], [33], [34], [35] establishing the existence of an approximation to any complex continuous function up to a desired level of accuracy (say $\varepsilon$) and generalization abilities of Neural Network. Main problem is related to the size of the hidden layer in attaining the desired accuracy. In 1993 Barron [27] gives a result on the number of neurons needed for the function approximation based on Fourier transform where each hidden neuron is activated with sigmoidal function by imposing certain conditions. Similar results were developed later in [28], [29], [30] related to the complexity of approximation and shows the importance of sigmoid activation functions.

\* Corresponding author. E-mail address: linupinto671@gmail.com

The approximation capabilities and generalization error that controlled by measures from statistical learning theory provides a good theoretical explanation on the expressive power of Neural Network. Sample complexity is a measure from statistical learning theory that relates with sample size and its distribution. The generalization capabilities of NN should be independent with the sample complexity after successful learning. This sample complexity is quantified based on a concept VC Dimension by Vapnik and Chervonenkis in 1971 [12]. The VC Dimension was later studied by Baum and Haussler in 1989; Maas in 1994, Sakurai in 1999 in [36],[37],[38] giving the lower and upper bounds for FNN with binary outputs and binary activation functions. Goldberg and Jerum in 1995 [41] analysed the bounds on VC Dimension with piecewise Polynomial activation functions. This was further improved by Koiran and Sontag in 1997 [39]. In 1997 Karpinski and Macintyre in [40] gives these bounds on more general activation functions like Tanh, logsig or atan. In 2009 Anthony and Barlett [29]gives a detailed explanation on VC Dimension with bounds on FNN with commonly used activation functions. All the related studies in the bounds points the theoretical explanation on the expressive power of Neural Network. Apart from VC Dimension concept there are several ways to analyze the expressive power of Neural network. Delallau and Bengio in 2011 [15] showed that for certain classes of polynomials sum-product networks can match approximately good with deep networks. In 2014 Montufar,et.al [16] showed that in network topology the number of linear regions could be estimated and increses exponentially in proportional to the number of layers. In 2014 Baimchini and Scarcelli [17] charecterises the topological properties of network functions by giving bounds on betti numbers. In 2015 and later in 2016 Telgarsky [18], [19] gives experimental results on the efficiency of deep Networks for classification problems. In 2015 Eldan and Shamir [20] proved that to approximate a function the dimension of the input space should be increased exponentially for a two layered feed forward network.

*1.2 Overview of the paper*

Section 2 first explains the network structure studied in the paper by giving a detailed explanation on basic notations and definitions related to single output classification problem with a single hidden layered Feed Forward Neural Network. Secondly it explains a new perspective taken in the development of the work that is the VC Dimension approach. The approach is systematically grooved with real algebraic definitions to fix the network size within a finite range. The results are supported with theorems. This section also explains the respective algorithm selected for the optimization and the proximity measures to evaluate the performance of the network framework proposed. Section 3 gives the experimentation results on the proposed work and section 4 gives a detailed discussion on the results with a suggestion on the required dimension on the sample space for the optimization with shallow networks. Section 4 gives the conclusion along with future study and limitations of the work.

## 2. METHOD

### 2.1 Background

This section summaries single layered feed forward neural network model used in the proposed study on a binary classification problem. This basic structure comprises three main constituents called layers that act as the processing tiers to form a regression curve for a mapping between the input vector and the targeted scalar. The first layer receive the inputs , the middle layer called the hidden layer activate the inputs with sigmoid activation functions to transform it into a more affluent domain by consuming connection weights and thresholds and the third layer called the output layer gives the corresponding output by the activation of linear threshold functions. The calculations in the hidden layer forms the underlying structure for the regression curve. This structure could further be tuned with weights and thresholds in the connections.

The notations relating the proposed network model consists of sigmoid activation functions $(\sigma_j)$ as the collection of functions in the hidden layer and $(\varphi)$ is the linear threshold activation using in the output layer. $w_{jk}^l$ denotes the weight that strengthens the connectivity between the $k^{th}$ neuron in the $l^{th}$ layer and $j^{th}$ neuron in the $l+1^{th}$ layer. $\vec{x}_i$ is the $i^{th}$ input vector $\vec{x}_i = (x_{1i}, x_{2i}, x_{3i}, \ldots \ldots x_{ni})$ for $i = 1,2,3,\ldots \ldots r$ samples collection E from the population $X$. Let S denotes a subset of E and each $S \subset E$ represents the state 'u' of the network. Hence there are a total of $2^r$ states for the network. $s_j^l$ denotes the weighted sum at $j^{th}$ node of $l^{th}$ layer with thresholder $t_j^{l-1}$. Weighted sum $s_j^l$ together with the threshold $t_j^{l-1}$ are activated with the activation function $f_j^l$. $W_T$ be the total weights and threshold in the network. $C_p$ be the total computational units in the network with $n$ input nodes and $m$ hidden nodes for the network shown in Fig. 1. Depth of the network is defined as the number of hidden layers in the network and width is defined as the number of nodes in each layer. For a single layered Feed forward network as shown in Fig. 1 depth is one and the width is the number of nodes $m$ in the hidden layer.

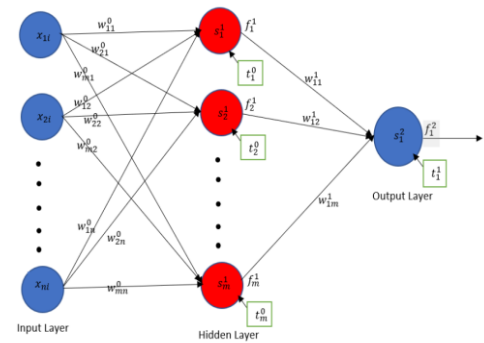

Fig. 1: Two Layered Feed Forward Sigmoid Network with Single Output

The functions represented by the processing state $'u'$ corresponding to each set $S \subset E \subset X$ is denoted as

$$f_u: X \times \Omega_w \to Y \qquad (1)$$

$f_u$ defined in (1) and shown in Fig. 1 is a well defined function such that

$$f_u(\bar{X}, w_u) = W^{1^T} \sigma(W^0 \bar{X} + T^0) + T^1 \qquad (2)$$

where $W^0 = \begin{bmatrix} w_{11}^0 & w_{12}^0 & \cdots & w_{1n}^0 \\ w_{21}^0 & w_{22}^0 & \cdots & w_{21}^0 \\ \cdots & \cdots & \cdots & \cdots \\ w_{m1}^0 & w_{m2}^0 & \cdots & w_{mn}^0 \end{bmatrix}$ is an $m \times n$ matrix,

$T_0 = \begin{bmatrix} t_1^0 \\ t_2^0 \\ \cdots \\ t_m^0 \end{bmatrix}$, $T^1 = [t_1^1]$, $W^{1^T} = \begin{bmatrix} w_{11}^1 \\ w_{12}^1 \\ \cdots \\ w_{1m}^1 \end{bmatrix}$, $\bar{X} \in R^n$, $\Omega_w$ collection of weight matrices $w_{u_{(n+3) \times m}}$ with each entry $w \in R$, $X \subset R^n$ and $Y \in [0,1]$. For a Two layered feedforward neural network as shown in Fig. 1 with activation functions fixed functionality of Neural Network $\mathcal{M}$ for a state $'u'$ could be represented in a $(n+3) \times m$ weight matrix $w_u \in \Omega_w$ as

$$w_u = \begin{bmatrix} w_{11}^0 & w_{12}^0 & \cdots & w_{1n}^0 & t_1^0 & w_{11}^1 & t_1^1 \\ w_{21}^0 & w_{22}^0 & \cdots & w_{2n}^0 & t_2^0 & w_{12}^1 & 0 \\ \cdots & \cdots & \cdots & \cdots & \cdots & \cdots & \cdots \\ w_{m1}^0 & w_{m2}^0 & \cdots & w_{mn}^0 & t_m^0 & w_{1m}^1 & 0 \end{bmatrix}_u \qquad (3)$$

## 2.2 VC Dimension based Approach

Literature review shows that single layered feed forward neural network with sigmoid activation functions has established its importance in the approximation of any measurable functions upto a desired level of accuracy. Main problem concerning this approximation is related to the size of the hidden layer. The concept of VC Dimension can address this problem to some extent but its probabilistic nature is limited to the computation of generalization error and is inappropriate in giving a higher judgement degree to identify a state 'u' approximately. Hence an improbabilistic definition in terms of continuity of functions designated as the $\varepsilon$-identifiable is applied to connect the concept with network size.

The function $f_u$ corresponding to the state 'u' as defined in (2) is said to be $\varepsilon$-identifiable on a set $S \subset E$, if for a predefined $\varepsilon$ value (considerably small and positive) there exist a $w_u \in \Omega_w$ such that

$$f_u(\bar{x}, w_u) = \begin{cases} \theta^1 \in B_\varepsilon(1) & \forall \bar{x} \in S \subset E \\ \theta^0 \in B_\varepsilon(0) & \forall \bar{x} \in S^C \subset E \end{cases} \qquad (4)$$

where $B_\varepsilon(1) = \{\theta^1 \in R : 1 - \varepsilon < \theta^1 \leq 1\}$, $B_\varepsilon(0) = \{\theta^0 \in R : 0 \leq \theta^0 < \varepsilon\}$ and $B_\varepsilon(1) \cap B_\varepsilon(0) = \emptyset$. Clearly the function denoted as $f_u^\varepsilon$ is well defined.

The function (2) is non-identifiable within an $\varepsilon$ range on a set $S \subset E$, if there exist no $w_u \in \Omega_w$ such that (4) satisfies.

The definition becomes more acceptable by giving a criteria to the chosen epsilon. If the chosen epsilon is least as possible then proximity could be well defined and the function turns out to be *least $\varepsilon$-identifiable function*.

The function $f_u$ corresponding to the state 'u' is said to be least $\varepsilon$-identifiable function on a set $S \subset E$, if for a predefined $\varepsilon$ value there exist a sequence of values $\varepsilon_0 < \varepsilon_1 < \cdots < \varepsilon_n = \varepsilon$ up to a desired level and $w_u \in \Omega_w$ such that

$$f_u(\bar{x}, w_u) = \begin{cases} B_{\varepsilon_0}(1) < B_{\varepsilon_1}(1) < \cdots < B_{\varepsilon_n}(1) & \forall \bar{x} \in S \subset E \\ B_{\varepsilon_0}(0) < B_{\varepsilon_1}(0) < \cdots < B_{\varepsilon_n}(0) & \forall \bar{x} \in S^C \subset E \end{cases} \qquad (5)$$

where $B_{\varepsilon_i}(1) = \{r \in R : 1 - \varepsilon_i < r \leq 1\}$ and $B_{\varepsilon_i}(0) = \{r \in R : 0 \leq r < \varepsilon_i\}$.

The advantage of this setting is that a state 'u' can be identified by many $\varepsilon$-identifiable functions whereas the least $\varepsilon$-identifiable function is one with lowest $\varepsilon$ value that could be determined and corresponding $w_u$ is the optimal weight matrix that could be identified with respect to the state 'u'.

The collection of least $\varepsilon$-identifiable functions for all $S \subset E \subset X$ computable by a neural network $\mathcal{M}$ forms a class on a sample set E called the least $\varepsilon$-identifiable function class and is denoted as $H_\mathcal{M}$. This act as the concept class for defining VC Dimension interms of $\varepsilon$ definitions.

$$H_\mathcal{M} = \left\{ f_u : \exists w_u \in \Omega_w \text{ and } f_u(\bar{x}, w_u) = \begin{cases} B_{\varepsilon_0}(1) & \forall \bar{x} \in S \subset E \\ B_{\varepsilon_0}(0) & \forall \bar{x} \in S^C \subset E \end{cases}, \forall S \subset E \subset X, \right\} \qquad (6)$$

*Index of* the $\varepsilon$-identifiable function class $H_\mathcal{M}$ *is the cardinality of the least $\varepsilon$-identifiable function class $H_\mathcal{M}$ denoted as $I(H_\mathcal{M})$.*

*The following result shows the properties of the index of $H_\mathcal{M}$ in identifying a state 'u' by the neural network $\mathcal{M}$.*

### Result 3.1:

*For a sample set E with cardinality 'r', Index of the $\varepsilon$-identifiable function class $H_\mathcal{M}$ has the following properties:*

i. $I(H_\mathcal{M}) \leq 2^r \; \forall r$

ii. If $I(H_\mathcal{M}) = 2^r$ then $\forall S \subset E \subset X$ there exist $f_u$ satisfying (11).

**Proof:**

If a state 'u' can be identified by neural network $\mathcal{M}$ then there exist $f_u$ corresponding to the state 'u'. For a binary classification problem there exist a total of $2^r$ states for a sample set E with cardinality 'r'. Hence maximum cardinality of *least $\varepsilon$-identifiable function class $H_\mathcal{M}$ is $2^r$.

Therefore $I(H_\mathcal{M}) \leq 2^r \; \forall r$.

If $I(H_\mathcal{M}) = 2^r$ then every state 'u' could be identified by $\mathcal{M}$. Hence there exist a least $\varepsilon$-identifiable function $f_u \; \forall S \subset E \subset X$.

For the sample population with cardinality $|X| = M$, $\mathcal{E}$ be the collection of subsets of X with cardinality $r < M$.

$$\mathcal{E} = \{E_p \subset X, |E_p| = r \; \forall p = 1,2, \dots M C_r\} \quad (7)$$

Vapnik and Chervonenkis, 1971, [12] proposed a general and standard tactic in depicting the complexity of a classification problem with respect to the sample space called VC Dimension. In terms of *Index of* the $\varepsilon$- identifiable function class $H_\mathcal{M}$ the concept of VC Dimension could be illustrated as the dimension of the $\varepsilon$-*identifiable function space* $\varepsilon L_\mathcal{M}$ and is the largest value of 'r' for which $I(H_\mathcal{M}) = 2^r$. It could be denoted as $D(\varepsilon L_\mathcal{M})$. The difference in this setting is that the application of $\varepsilon$ concept brings more optimal approximation to the concept of VC Dimension.

Hence mathematically expressed as

$$D(\varepsilon L_\mathcal{M}) = \max\{r : I(H_\mathcal{M}) = 2^r \text{ for some } E_p \in \mathcal{E}\} \quad (8)$$

Certainty of Neural Network $\mathcal{M}$ in identifying a function $f_u$ corresponding to a state with respect to sample set E lies in the concept of the dimension of $\varepsilon$-*identifiable function space* $\varepsilon L_\mathcal{M}$ can be proved by the following theorem.

**Theorem 3.2**

*For a sample set E with cardinality r on an n dimensional space $X = R^n$, if the dimension of the $\varepsilon$-identifiable function space $\varepsilon L_\mathcal{M}$ of Neural Network $\mathcal{M}$ is $D(\varepsilon L_\mathcal{M}) = r$ then for each state 'u' corresponding to a sample set E there exist a function $f_u^\varepsilon$ and weight matrix $w_u \in \Omega_w$ such that*

$$f_u(\bar{x}, w_u) = \begin{cases} B_\varepsilon(1) & \forall \bar{x} \in S \subset E \\ B_\varepsilon(0) & \forall \bar{x} \in S^C \subset E \end{cases} \quad (9)$$

**Proof:**

If $D(\varepsilon L_\mathcal{M}) = r$ then by Sec F, $P_\mathcal{M}(r) = 2^r$. This implies $\max_{E_p \in \mathcal{E}} \{|H_\mathcal{M}|\} = 2^r$ by (11). This implies $|H_\mathcal{M}| = 2^r$ for some $E_p \subset X$ with $|E_p| = r$. Then by (10) there exist some $w_u \in \Omega_w$ and hence a function $f_u^\varepsilon$ for each $S \subset E \subset X$ such that

$$f_u(\bar{x}, w_u) = \begin{cases} B_\varepsilon(1) & \forall \bar{x} \in S \subset E \\ B_\varepsilon(0) & \forall \bar{x} \in S^C \subset E \end{cases}$$

### 2.3 Structuring Hidden Layers in $\varepsilon$-identifiable function space

VC Dimension could give an explicit structure to the width of a single layered feed forward sigmoid network. The section deals with determining the bounds on the width of the network corresponding to the VC bounds on generalization error. The lower bound on the VC Dimension of the proposed network was selected based on the result of Koiran and Sontag (1997), [39] which shows that the lower bound of the VC Dimension for Linear Threshold Network $\mathcal{L}$ is also the lower bound of the VC Dimension for sigmoid Network $\mathcal{M}$.

Sakurai (1999), [38] gives lower bound for $\mathcal{L}$ with $n \geq 3$ and $C_p \leq 2^{\frac{n}{2}-2}$. For real inputs this lower bound is given by

$$D(\varepsilon L_\mathcal{L}) \geq \frac{nC_p}{8} \log_2 \frac{C_p}{4} \quad (10)$$

Combining the studies of Macintyre and Sontag (1993), [43], Karpinski and Macintyre (1997), [40] and Goldberg and Jerrum (1995), [41] the upper bound on the VC Dimension for a feed forward Neural Network $\mathcal{M}$ with $W_T$ parameters, $C_p$ computational units, linear threshold unit in the output layer and sigmoid functions in the hidden layer is fixed as

$$D(\varepsilon L_\mathcal{M}) \leq (W_T C_p)^2 + 11 W_T C_p \log_2(18 W_T C_p^2) \quad (11)$$

With the results obtained from the literature in (10) and (11) along with the explicit definition of VC Dimension as the dimension of the $\varepsilon$-*identifiable function space* $\varepsilon L_\mathcal{M}$, following inequality on the bounds on VC Dimension is obtained which is valid for a binary classification problem with real inputs that uses feed forward Neural Network topology with sigmoid activation functions in the hidden layer and linear function in the output layer.

$$\frac{nC_p}{8} \log_2 \frac{C_p}{4} \leq D(\varepsilon L_\mathcal{L}) \leq D(\varepsilon L_\mathcal{M}) \leq (W_T C_p)^2 + 11 W_T C_p \log_2(18 W_T C_p^2) \quad (12)$$

$W_T$, total weights and threshold in the network and $C_p$, total computational units in the network depends on $n$ input nodes and $m$ hidden nodes

$$W_T = nm + 2m + 1 \quad (13)$$
$$C_p = m + 1 \quad (14)$$

The next approach is to find the approximate bounds on the width of the network based on the inequality obtained in (12). Before finding the approximate bounds on the width of the network it is important to analyses whether such bounds exists and finite. The following theorems give a detailed proof on this regard.

**Theorem 3.3**

*For a nonlinear classification with a two layered feedforward neural network $\mathcal{M}$ as shown in Fig.1 with total weights and threshold $W_T$, total computational units $C_p$, n input nodes and m hidden nodes there exist a lower bound $l_m \in R^+$ over the width of the network 'm' such that (12) holds and satisfies the condition $C_p \leq 2^{\frac{n}{2}-2} \; \forall \; n > 4$.*

**Proof:**

By Result 3.1 of Sec 2.2, the certainty of Neural Network $\mathcal{M}$ in identifying a function corresponding to a state 'u' for a sample set with cardinality r is given by $D(\varepsilon L_\mathcal{L}) \leq r \leq D(\varepsilon L_\mathcal{M})$. Applying (13) and (14) in (12) implies

$$\frac{n(m+1)}{8} \log_2 \frac{(m+1)}{4} \leq r \leq Q(m)[Q(m) + 11 \log_2(18(m+1)Q(m))] \quad (15)$$

Q(m) is a quadratic polynomial of the form
$$Q(m) = (n+2)m^2 + (n+3)m + 1 \quad (16)$$

Consider $r \leq Q(m)[Q(m) + 11 \log_2(18(m+1)Q(m))]$
implies
$r \leq Q(m)[Q(m) + 11 \log_2(18Q(m)) + 11 \log_2(m+1)]$
Applying the condition $C_p = (m+1) \leq 2^{\frac{n}{2}-2}$
$r \leq Q(m)[Q(m) + 11 \log_2(18Q(m)) + 11(\frac{n}{2}-2)]$
implies $r \leq Q(m)[Q(m) + 11 \times 18\ Q(m) + 11(\frac{n}{2}-2)]$
implies $199\ Q(m)^2 + 11\left(\frac{n}{2}-2\right)Q(m) - r \geq 0$ (17)
Consider $199\ Q(m)^2 + 11\left(\frac{n}{2}-2\right)Q(m) - r = 0$ (18)
Now Discriminant,
$D = [11\left(\frac{n}{2}-2\right)]^2 + 4 \times 199\ r > 0, \forall r > 0, n > 4$ (19)
Suppose $|\sqrt{D}| < 11\left(\frac{n}{2}-2\right)$ then
$Q(m) = \frac{-11(n/2-2)+\sqrt{(D)}}{2 \times 199} < 0$ (20)
is a contradiction since by (16) $Q(m)$ is positive. Hence $|\sqrt{D}| > 11\left(\frac{n}{2}-2\right)$ and there exist a positive real root $\beta$ for (18) $\forall r > 0, n > 4$. Hence (17) implies
$Q(m) \geq \beta$ (21)
Since $Q(m) > 1\ \forall n > 4, m > 0$
Choose a real number $\gamma$ such that $Q(m) \geq \gamma > \beta$ and $\gamma > 1$.
Consider $Q(m) \geq \gamma$ (22)
Implies $(n+2)m^2 + (n+3)m + 1 - \gamma \geq 0$ (23)
$(n+2)m^2 + (n+3)m + 1 - \gamma = 0$ (24)
Discriminant of (23) is given by
$d = (n+3)^2 + 4(n+2)(\gamma-1) > 0$
Hence the roots are real and
$m = \frac{-(n+3) \pm |\sqrt{d}|}{2(n+2)}$
Suppose $|\sqrt{d}| < n+3$
Implies $\sqrt{(n+3)^2 + 4(n+2)(\gamma-1)} < n+3$, contradiction to the assumption $\gamma > 1$. Hence $|\sqrt{d}| > n+3$ and there exist a positive real root $l_m$ on the solution set of (24) and is a lower bound for the solution satisfying (23) and hence (17).

### Theorem 3.4

*For a nonlinear classification with a two layered feedforward neural network $\mathcal{M}$ as shown in Fig.1 with total weights and threshold $W_T$, total computational units $C_p$, n input nodes and m hidden nodes there exist an upper bound $L_m \in R^+$ over the width of the network 'm' with respect to the size r and dimension 'n' of sample space such that (12) holds and*

$$L_m = mini\{k_1, k_2\}$$

*where $k_1 = \frac{16r}{n(n-8)} - 1, k_2 = 2^{\frac{n}{2}-2} - 1$ and $C_p \leq 2^{\frac{n}{2}-2}\ \forall\ n > 8$. Also the sufficient condition for the existence of least upper bound $L_m^* \in R^+$ is given by $r = 2^{\frac{n}{2}-6}\ n(n-8)$.*

### Proof:

By (15) $\frac{n(m+1)}{8} \log_2 \frac{(m+1)}{4} \leq r$ (25)
Also $m + 1 \leq 2^{\frac{n}{2}-2}\ \forall\ n > 8$ (26)
Hence there exist a $N \in Z^+$ such that (27)
$n2^{\frac{n}{2}-2}\left(\frac{n}{2}-4\right) \leq 8r, \forall n \leq N$
implies
$2^{\frac{n}{2}-2} - 1 \leq \frac{16r}{n(n-8)} - 1\ \forall n \leq N$ (28)
implies $m \leq \frac{16r}{n(n-8)} - 1\ \forall n \leq N$ (29)

From (26) and (29)
$L_m = mini\left\{\frac{16r}{n(n-8)} - 1, 2^{\frac{n}{2}-2} - 1\right\} \forall n \leq N, n > 8$ (30)

If (26) and (29) coincides we get a least upper bound for the width of the network and is given by
$r = 2^{\frac{n}{2}-6} n(n-8), n > 8$ (31)
is a sufficient condition for the existence of least upper bound $L_m^* \in R^+$ on the width of the network with n input nodes and r samples.
Based on the conditions on theorem 3.3 and theorem 3.4 the bounds on the width of the network could be structured.

### Result 3.5

The width of a single layered feed forward sigmoid network should be $l_m < m < L_m$ such that $C_p \leq 2^{\frac{n}{2}-2}\ \forall\ n > 8$ where $l_m$ is the least width possible that satisfies the relation $199\ Q(m)^2 + 11\left(\frac{n}{2}-2\right)Q(m) - r \geq 0$, $Q(m) = (n+2)m^2 + (n+3)m + 1$ and $L_m$ is the highest width possible satisfying $L_m = mini\left\{\frac{16r}{n(n-8)} - 1, 2^{\frac{n}{2}-2} - 1\right\} \forall n \leq N$ for some $N \in Z^+$.

Based on this result the nodes in the hidden layers are structured in between $l_m$ and $L_m$ reducing the unnecessary computations.

### 2.4 Data processing with back propagation algorithm

After fixing the network topology as explained in section 2.3 the optimization is carried out with backpropagation algorithm. Backpropagation gives a thorough perceptions on the overall behavior corresponding to the change of parameters of the network. The hidden layer activation function is log sigmoid and the output activation function is linear. Hence the function representing the forward pass of a fully connected single hidden layered network with an output node could be represented by (2). Performance of the algorithm is based on a measure called error function, mathematically expressed as
$E(w_u) = \frac{1}{2r} \sum_{\bar{x}} \|y_u(\bar{x}) - f_u(w_u, \bar{x})\|^2$ (32)

Since there is only a single output node (32) could be further represented by
$E(w_u) = \frac{1}{2r} \sum_{\bar{x}} (y_u(\bar{x}) - \sum_{k=1}^{m} w_{1k}^1 \sigma(z_k^1) + t_1^1)^2$ (33)
Here $y_u(\bar{x})$ denotes the targeted output and the summation is taken over r samples. The data is processed by backpropagation algorithm to determine the optimal weights that reduces the error based on the error function using a gradient descent method. The success of backpropagation algorithm depends on finding the direction to which the error function attains its global minimum.

The gradient of error with respect to the weights between the layers is given by

$$\frac{\partial E}{\partial w_{1k}^1} = -(1+e^{-\sum_{j=1}^n w_{kj}^0 \overline{x_l}+t_k^0})^{-1} \qquad (34)$$

$$\frac{\partial E}{\partial t_1^1} = -1 \qquad (35)$$

$$\frac{\partial E}{\partial w_{kj}^0} = -w_{1k}^1 \frac{e^{-\sum_{j=1}^n w_{kj}^0 \overline{x_l}+t_k^0}}{(1+e^{-\sum_{j=1}^n w_{kj}^0 \overline{x_l}+t_k^0})^2} x_{ji} \qquad (36)$$

$$\frac{\partial E}{\partial t_k^0} = -w_{1k}^1 \frac{e^{-\sum_{j=1}^n w_{kj}^0 \overline{x_l}+t_k^0}}{(1+e^{-\sum_{j=1}^n w_{kj}^0 \overline{x_l}+t_k^0})^2} \qquad (37)$$

The gradient of errors is used to update the weight of the network by gradient descent algorithm

$$w_{1k}^1 = w_{1k}^1 - \frac{\eta}{r}\sum_{\overline{x_l}}\frac{\partial E}{\partial w_{1k}^1} \qquad (38)$$

$$t_1^1 = t_1^1 - \frac{\eta}{r}\sum_{\overline{x_l}}\frac{\partial E}{\partial t_1^1} \qquad (39)$$

$$w_{kj}^0 = w_{kj}^0 - \frac{\eta}{r}\sum_{\overline{x_l}}\frac{\partial E}{\partial w_{kj}^0} \qquad (40)$$

$$t_k^0 = t_k^0 - \frac{\eta}{r}\sum_{\overline{x_l}}\frac{\partial E}{\partial t_k^0} \qquad (41)$$

where $\eta$ is the parameter that represents the learning rate.

## 2.5 Dataset

The Network structured in section 2.3 is experimentally studied with three datasets that includes types of glass dataset, thyroid dataset and wine vintage dataset. The scope of the proposed study is limited to binary classification and hence the target attributes are designed accordingly. Description on each dataset and its attributes used for the study is shown in table1

TABLE 1
DESCRIPTION OF DATASETS UNDER STUDY

| Dataset | Data Description | Data Inputs/Attributes | Data Targets | Cardinality of the Sample set |
|---|---|---|---|---|
| Types of Glass | Classifies Glasses based on glass Chemistry | Refractive Index, Sodium, Magnesium, Aluminum, Silicon, Potassium, Calcium, Barium, Iron | Window Glass, Non Window Glass | 214 |
| Thyroid Dataset | Classify patients according to thyroid functioning based on clinical records | There are 21 data inputs based on clinical records with 15 binary and 6 continuous patient attributes | Normal, Abnormal | 7200 |
| Wine Vintage | Classifies wines from three winerys in Italy based on constituents found through chemical analysis | Alcohol, Malic Acid, Ash, Alcalinity of Ash, Magnesium, Total phenols, Flavanoids, Non Flavanoid Phenols, Color intensity, Hue, OD280/OD315 of diluted wines, Proline | Vineyard 1, Others | 178 |

*Source:* UCI Machine learning Repository (http://mlearn.ics.uci.edu/MLRepository.html)

### 2.6 Performance Evaluation and Measures on Proximity of a Network $\mathcal{M}$

With the data sets explained in section 2.5 and the optimal algorithm decided as per section 2.4, the network structured in section 2.3 is evaluated experimentally with the measures on proximity. This paper analyses the proximity of the network $\mathcal{M}$ based on two measures, Dissimilarity Measure and Mean Squared Error.

Let $g_{1i}, g_{2i}, g_{3i}$ be a real valued function defined on discrete values $i = 1,2,3,\ldots M$. Then

$$d_{\mathcal{M}}(g_{1i}, g_{2i}, g_{3i}) = \frac{1}{100}\left[\sum_{i=1}^{M}\sum_{\substack{j=1\\j\neq k}}^{3}|g_{ji} - g_{ki}|\right] \qquad (42)$$

is the dissimilarity measure on a network $\mathcal{M}$ if the following conditions hold.

i.   $d_{\mathcal{M}}(g_{1i}, g_{2i}, g_{3i}) \geq 0$

ii.   $d_{\mathcal{M}}(g_{1i}, g_{2i}, g_{3i}) = 0$ iff $g_{1i} = g_{2i} = g_{3i}$

iii.  $d_{\mathcal{M}}(g_{1i}, g_{2i}, g_{3i}) = d_{\mathcal{M}}(g_{2i}, g_{1i}, g_{3i}) = d_{\mathcal{M}}(g_{1i}, g_{3i}, g_{2i}) = d_{\mathcal{M}}(g_{3i}, g_{2i}, g_{1i}) = d_{\mathcal{M}}(g_{2i}, g_{3i}, g_{1i}) = d_{\mathcal{M}}(g_{3i}, g_{1i}, g_{2i})$

Let $h_i, t_i$ be real valued functions defined on discrete values $i = 1,2,3,....r$. then Mean Squared Error on a network $\mathcal{M}$ is defined as

$$E_r(\mathcal{M}) = \sum_{i=1}^{r}(h_i - t_i)^2 \qquad (43)$$

## 3. RESULTS

At first suitable attribute dimension for a given sample size was studied on each dataset based on the theoretical analysis as per theorem 3.4 and the results are plotted on table 2. Application of theorem 3.3 gives the lower bound on the dimension of hidden layer and corresponding upper bound is obtained from table 2 for each dataset. The results are plotted separately on table 3.

TABLE 2
TABULAR RESULTS ON THEOREM 3.4

| | Types of Glass Dataset (r=214) | | | Thyroid Dataset (r=7200) | | | Wine Vintage Dataset (r=178) | | |
|---|---|---|---|---|---|---|---|---|---|
| Attribute size (n) | k1 | k2 | $L_m$ | Attribute size (n) | k1 | k2 | Attribute size (n) | k1 | k2 |
| 9 | 379.4444444 | 4.656854249 | 4.656854 | 9 | 12799 | 4.656854 | 4.656854 | 9 | 315.44444 | 4.656854249 | 4.656854 |
| 10 | 170.2 | 7 | 7 | 10 | 5759 | 7 | 7 | 10 | 141.4 | 7 | 7 |
| 11 | 102.7575758 | 10.3137085 | 10.31371 | 11 | 3489.9091 | 10.31371 | 10.31371 | 11 | 85.30303 | 10.3137085 | 10.31371 |
| 12 | 70.33333333 | 15 | 15 | 12 | 2399 | 15 | 15 | 12 | 58.333333 | 15 | 15 |
| 13 | 51.67692308 | 21.627417 | 21.62742 | 13 | 1771.3077 | 21.62742 | 21.62742 | 13 | 42.815385 | 21.627417 | 21.62742 |
| 14 | 39.76190476 | 31 | 31 | 14 | 1370.4286 | 31 | 31 | 14 | 32.904762 | 31 | 31 |
| 15 | 31.60952381 | 44.254834 | 31.60952 | 15 | 1096.1429 | 44.25483 | 44.25483 | 15 | 26.12381 | 44.254834 | 26.12381 |
| 16 | 25.75 | 63 | 25.75 | 16 | 899 | 63 | 63 | 16 | 21.25 | 63 | 21.25 |
| 17 | 21.37908497 | 89.50966799 | 21.37908 | 17 | 751.94118 | 89.50967 | 89.50967 | 17 | 17.614379 | 89.50966799 | 17.61438 |
| 18 | 18.02222222 | 127 | 18.02222 | 18 | 639 | 127 | 127 | 18 | 14.822222 | 127 | 14.82222 |
| 19 | 15.38277512 | 180.019336 | 15.38278 | 19 | 550.19617 | 180.0193 | 180.0193 | 19 | 12.626794 | 180.019336 | 12.62679 |
| 20 | 13.26666667 | 255 | 13.26667 | 20 | 479 | 255 | 255 | 20 | 10.866667 | 255 | 10.86667 |
| 21 | 11.54212454 | 361.038672 | 11.54212 | 21 | 420.97802 | 361.0387 | 361.0387 | 21 | 9.4322344 | 361.038672 | 9.432234 |
| 22 | 10.11688312 | 511 | 10.11688 | 22 | 373.02597 | 511 | 511 | 22 | 8.2467532 | 511 | 8.246753 |
| 23 | 8.924637681 | 723.0773439 | 8.924638 | 23 | 332.91304 | 723.0773 | 332.913 | 23 | 7.2550725 | 723.0773439 | 7.255072 |
| 24 | 7.916666667 | 1023 | 7.916667 | 24 | 299 | 1023 | 299 | 24 | 6.4166667 | 1023 | 6.416667 |
| 25 | 7.056470588 | 1447.154688 | 7.056471 | 25 | 270.05882 | 1447.155 | 270.0588 | 25 | 5.7011765 | 1447.154688 | 5.701176 |
| 26 | 6.316239316 | 2047 | 6.316239 | 26 | 245.15385 | 2047 | 245.1538 | 26 | 5.0854701 | 2047 | 5.08547 |
| 27 | 5.674463938 | 2895.309376 | 5.674464 | 27 | 223.5614 | 2895.309 | 223.5614 | 27 | 4.5516569 | 2895.309376 | 4.551657 |
| 28 | 5.114285714 | 4095 | 5.114286 | 28 | 204.71429 | 4095 | 204.7143 | 28 | 4.0857143 | 4095 | 4.085714 |
| 29 | 4.622331691 | 5791.618751 | 4.622332 | 29 | 188.16256 | 5791.619 | 188.1626 | 29 | 3.6765189 | 5791.618751 | 3.676519 |
| 30 | 4.187878788 | 8191 | 4.187879 | 30 | 173.54545 | 8191 | 173.5455 | 30 | 3.3151515 | 8191 | 3.315152 |
| 31 | 3.802244039 | 11584.2375 | 3.802244 | 31 | 160.57083 | 11584.24 | 160.5708 | 31 | 2.9943899 | 11584.2375 | 2.99439 |
| 32 | 3.458333333 | 16383 | 3.458333 | 32 | 149 | 16383 | 149 | 32 | 2.7083333 | 16383 | 2.708333 |
| 33 | 3.15030303 | 23169.47501 | 3.150303 | 33 | 138.63636 | 23169.48 | 138.6364 | 33 | 2.4521212 | 23169.47501 | 2.452121 |
| 34 | 2.873303167 | 32767 | 2.873303 | 34 | 129.31674 | 32767 | 129.3167 | 34 | 2.2211195 | 32767 | 2.221719 |
| 35 | 2.623280423 | 46339.95001 | 2.62328 | 35 | 120.90476 | 46339.95 | 120.9048 | 35 | 2.0137566 | 46339.95001 | 2.013757 |
| 36 | 2.396825397 | 65535 | 2.396825 | 36 | 113.28571 | 65535 | 113.2857 | 36 | 1.8253968 | 65535 | 1.825397 |
| 37 | 2.191053122 | 92680.90002 | 2.191053 | 37 | 106.36253 | 92680.9 | 106.3625 | 37 | 1.6542404 | 92680.90002 | 1.65424 |
| 38 | 2.003508772 | 131071 | 2.003509 | 38 | 100.05263 | 131071 | 100.0526 | 38 | 1.4982456 | 131071 | 1.498246 |
| 39 | 1.832092639 | 185362.8 | 1.832093 | 39 | 94.28536 | 185362.8 | 94.28536 | 39 | 1.3556658 | 185362.8 | 1.355666 |
| 40 | 1.675 | 262143 | 1.675 | 40 | 89 | 262143 | 89 | 40 | 1.225 | 262143 | 1.225 |
| 41 | 1.530672579 | 370726.6001 | 1.530673 | 41 | 84.144124 | 370726.6 | 84.14412 | 41 | 1.104952 | 370726.6001 | 1.104952 |
| 42 | 1.397759104 | 524287 | 1.397759 | 42 | 79.672269 | 524287 | 79.67227 | 42 | 0.9943978 | 524287 | 0.994398 |
| 43 | 1.275083056 | 741454.2002 | 1.275083 | 43 | 75.54485 | 741454.2 | 75.54485 | 43 | 0.8923588 | 741454.2002 | 0.892359 |
| 44 | 1.161616162 | 1048575 | 1.161616 | 44 | 71.727273 | 1048575 | 71.72727 | 44 | 0.7979798 | 1048575 | 0.79798 |
| 45 | 1.056456456 | 1482909.4 | 1.056456 | 45 | 68.189189 | 1482909 | 68.18919 | 45 | 0.7105105 | 1482909.4 | 0.710511 |
| 46 | 0.95881069 | 2097151 | 0.95881 | 46 | 64.90389 | 2097151 | 64.90389 | 46 | 0.6292906 | 2097151 | 0.629291 |
| 47 | 0.867975996 | 2965819.801 | 0.867976 | 47 | 61.847791 | 2965820 | 61.84779 | 47 | 0.553737 | 2965819.801 | 0.553737 |
| 48 | 0.783333333 | 4194303 | 0.783333 | 48 | 59 | 4194303 | 59 | 48 | 0.4833333 | 4194303 | 0.483333 |
| 49 | 0.704330513 | 5931640.602 | 0.704331 | 49 | 56.341961 | 5931641 | 56.34196 | 49 | 0.4176207 | 5931640.602 | 0.417621 |
| 50 | 0.63047619 | 8388607 | 0.630476 | 50 | 53.857143 | 8388607 | 53.85714 | 50 | 0.3561905 | 8388607 | 0.35619 |
| 51 | 0.561331509 | 11863282.2 | 0.561332 | 51 | 51.53078 | 11863282 | 51.53078 | 51 | 0.2986776 | 11863282.2 | 0.298678 |
| 52 | 0.496503497 | 16777215 | 0.496503 | 52 | 49.34965 | 16777215 | 49.34965 | 52 | 0.2447552 | 16777215 | 0.244755 |
| 53 | 0.435639413 | 23726565.41 | 0.435639 | 53 | 47.301887 | 23726565 | 47.30189 | 53 | 0.19413 | 23726565.41 | 0.19413 |
| 54 | 0.3784219 | 33554431 | 0.378422 | 54 | 45.376812 | 33554431 | 45.37681 | 54 | 0.1465378 | 33554431 | 0.146538 |
| 55 | 0.324564797 | 47453131.81 | 0.324565 | 55 | 43.564797 | 47453132 | 43.5648 | 55 | 0.1017408 | 47453131.81 | 0.101741 |
| 56 | 0.273809524 | 67108863 | 0.27381 | 56 | 41.857143 | 67108863 | 41.85714 | 56 | 0.0595238 | 67108863 | 0.059524 |
| 57 | 0.225921948 | 94906264.62 | 0.225922 | 57 | 40.245972 | 94906265 | 40.24597 | 57 | 0.0196921 | 94906264.62 | 0.019692 |
| 58 | 0.180689655 | 134217727 | 0.18069 | 58 | 38.724138 | 1.34E+08 | 38.72414 | 58 | -0.017931 | 134217727 | -0.01793 |
| 59 | 0.137919575 | 189812530.2 | 0.13792 | 59 | 37.285145 | 1.9E+08 | 37.28514 | 59 | -0.0535061 | 189812530.2 | -0.05351 |
| 60 | 0.097435897 | 268435455 | 0.097436 | 60 | 35.923077 | 2.68E+08 | 35.92308 | 60 | -0.0857175 | 268435455 | -0.08718 |
| 61 | 0.059078255 | 379625061.5 | 0.059078 | 61 | 34.632539 | 3.8E+08 | 34.63254 | 61 | -0.1190844 | 379625061.5 | -0.11908 |
| 62 | 0.02270019 | 536870911 | 0.0227 | 62 | 33.408602 | 5.37E+08 | 33.4086 | 62 | -0.1493429 | 536870911 | -0.14934 |
| 63 | -0.011832612 | 759250124 | -0.01183 | 63 | 32.246753 | 7.59E+08 | 32.24675 | 63 | -0.1780664 | 759250124 | -0.17807 |
| 64 | -0.044642857 | 1073741823 | -0.04464 | 64 | 31.142857 | 1.07E+09 | 31.14286 | 64 | -0.2053571 | 1073741823 | -0.20536 |
| 65 | -0.075843455 | 1518500249 | -0.07584 | 65 | 30.093117 | 1.52E+09 | 30.09312 | 65 | -0.231309 | 1518500249 | -0.23131 |

TABLE 3
NETWORK WIDTH RANGE ON EACH DATASET

| Dataset | Lower Bound On Network Width | Upper Bound On Network Width |
|---|---|---|
| Types of Glass | 1 | 5 |
| Thyroid | 1 | 361 |
| Wine Vintage | 1 | 21 |

Next study was done experimentally with nntool box on Mathlab 2018 (b) software and the Network is trained by changing the hidden layer size within and outside the range specified in table 3 to see the best performance plot. In the case of thyroid dataset as the upper bound is large and network training till that range is computationally complex, the best performance is plotted upto a possible hidden layer size. The network training is done till the mean squared error is minimum and the training testing and the validation results obtained with the number of epochs in the x-axis and the mean squared error corresponding to each epoch against the y-axis coincides. The proximity of the graphs is measured based on a measure called measure of proximity of network $\mathcal{M}$ or Dissimilarity measure ($d_{\mathcal{M}}$) as defined in Sec.2.6. The Optimal Network is selected with Minimum $d_{\mathcal{M}}$ and least Mean squared error. The best training performance on each width size is recorded in table 4 and its analysis is done by plotting bar graph (Fig 2, Fig 3, Fig 4) corresponding to the network width against the performance measures for three different datasets.

TABLE 4
PERFORMANCE EVALUATION ON EACH DATASET

| Dataset | Attribute | Network Width | MSE | Dissimilarity Measure | Network Width | MSE | Dissimilarity Measure | Network Width | MSE | Dissimilarity Measure |
|---|---|---|---|---|---|---|---|---|---|---|
| Types of Glass | 9 | 1 | 0.072046 | 0.034059 | 4 | 0.18393 | 0.161554 | 7 | 0.081343 | 0.404684 |
| | | 2 | 0.18606 | 0.15373 | 5 | 0.18736 | 0.132167 | 8 | 0.10097 | 0.17574 |
| | | 3 | 0.11571 | 0.491896 | 6 | 0.076018 | 0.184104 | 9 | 0.089524 | 0.266313 |
| Thyroid Dataset | 21 | 1 | 0.022802 | 0.012865 | 15 | 0.02713 | 0.018116 | 62 | 0.041717 | 0.120191 |
| | | 2 | 0.022306 | 0.035519 | 18 | 0.027175 | 0.028797 | 65 | 0.047965 | 0.038426 |
| | | 3 | 0.02755 | 0.007137 | 21 | 0.034139 | 0.12296 | 69 | 0.044806 | 0.048954 |
| | | 4 | 0.021166 | 0.038158 | 27 | 0.028771 | 0.039152 | 70 | 0.038365 | 0.113186 |
| | | 5 | 0.029979 | 0.023980 | 32 | 0.030619 | 0.081230 | 72 | 0.0407 | 0.062029 |
| | | 6 | 0.02959 | 0.009491 | 36 | 0.037501 | 0.110415 | 75 | 0.034194 | 0.104610 |
| | | 9 | 0.031908 | 0.040902 | 44 | 0.037868 | 0.062041 | 81 | 0.041241 | 0.064936 |
| | | 11 | 0.026016 | 0.057034 | 50 | 0.039362 | 0.145853 | 95 | 0.041506 | 0.082120 |
| | | 12 | 0.028535 | 0.055387 | 57 | 0.03251 | 0.125610 | 99 | 0.04137 | 0.062851 |
| Wine Vintage Dataset | 13 | 1 | 0.18412 | 0.168229 | 8 | 0.075268 | 0.218679 | 18 | 0.089226 | 0.450785 |
| | | 2 | 0.22897 | 0.111716 | 10 | 0.14376 | 0.141477 | 19 | 0.10944 | 0.473913 |
| | | 3 | 0.21279 | 0.257802 | 11 | 0.099735 | 0.231491 | 20 | 0.072471 | 0.400476 |
| | | 4 | 0.17069 | 0.178478 | 13 | 0.079918 | 0.194489 | 21 | 0.088083 | 0.359499 |
| | | 5 | 0.20205 | 0.17549 | 15 | 0.078167 | 0.916329 | 24 | 0.089768 | 0.173866 |
| | | 6 | 0.10006 | 0.257095 | 16 | 0.066525 | 0.301176 | 32 | 0.033028 | 0.448941 |
| | | 7 | 0.091456 | 0.125131 | 17 | 0.087048 | 0.380771 | 45 | 0.048058 | 1.007583 |

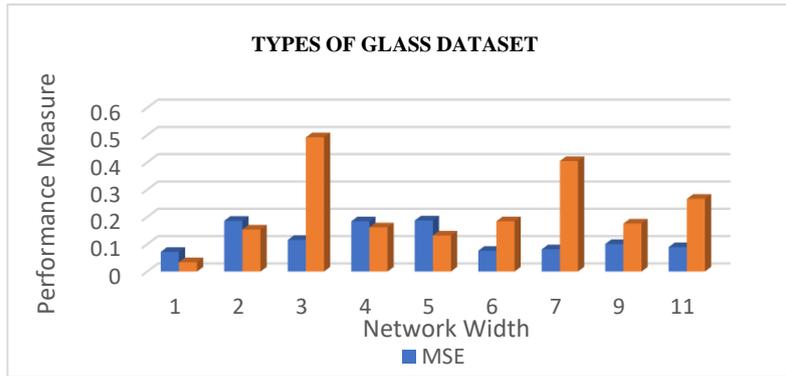

Fig 2: Performance of Neural network on Types of Glass Dataset

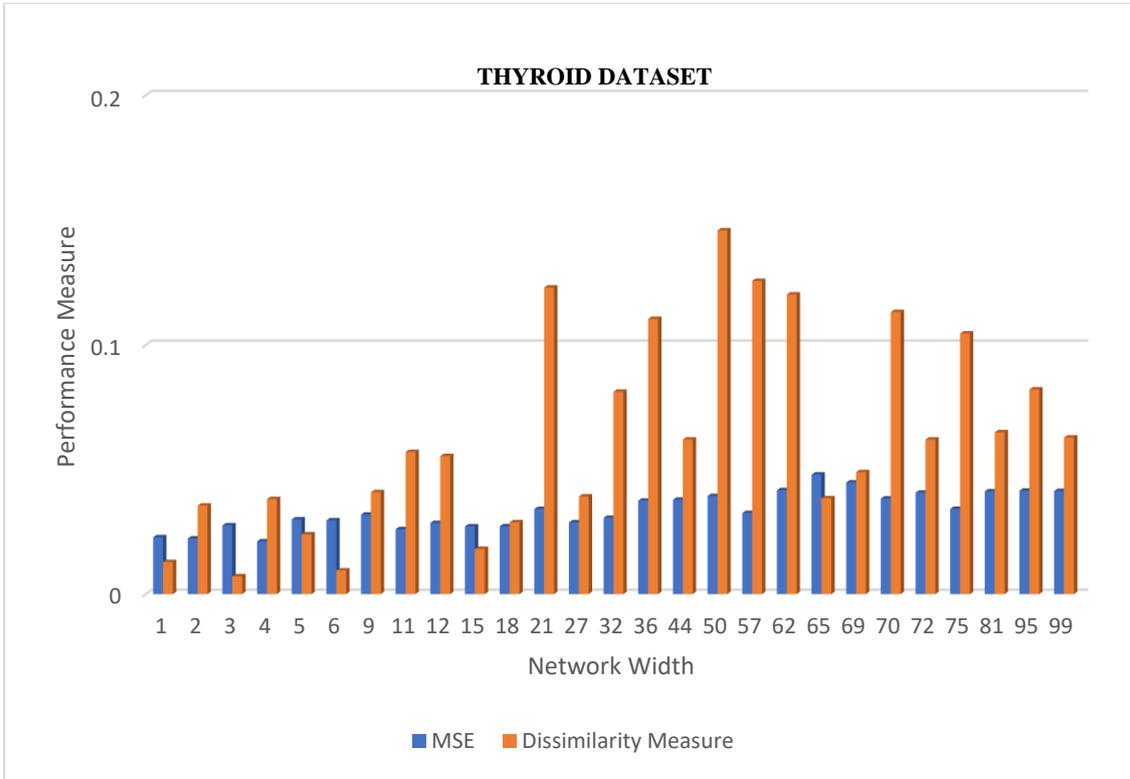

Fig 3: Performance of Neural network on Thyroid Dataset

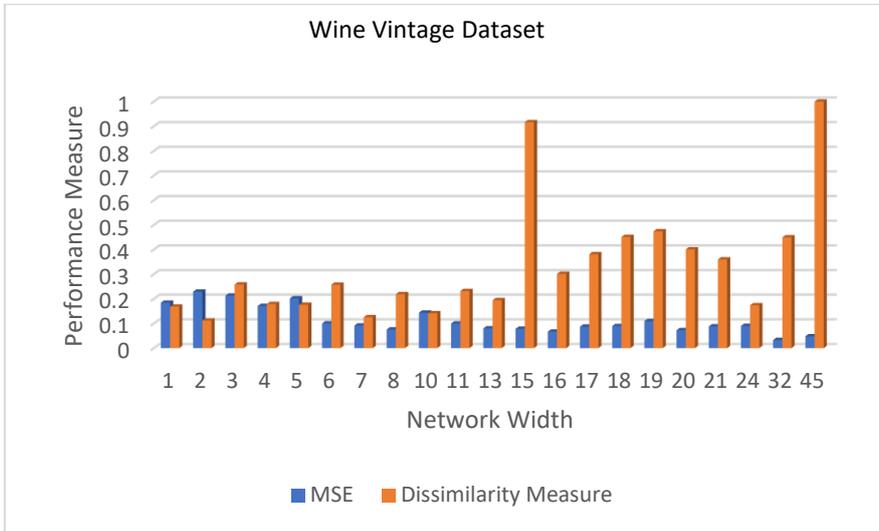

Fig 4: Performance of Neural network on Wine Vintage Dataset

## 4. DISCUSSION

The studies done by [8],[9],[10] related to network size do not provide an explicit result on the size of the network and assume unlimited number of network units. Many common networks with a single hidden layer have universal approximation property and can perform highly well than deep ones. The theoretical (Table 2) as well as the experimental results (Fig 2,3,4) in this paper shows the approximate width is proportional to the dimension of dataset.

The condition $C_p \leq 2^{\frac{n}{2}-2} \; \forall \; n > 8$ implies the appropriate attribute size is when $k_1 < k_2$. The upper bound $L_m$ on the dimension of the hidden layer is $L_m = mini\{k_1, k_2\}$. By the application of theorem 3.4 minimum number of nodes in the hidden layer for each data set is at least one. As per the results obtained on table 2 the attribute size and the suggested range on the dimension of hidden layer for each dataset is shown in table 5. Types of Glass dataset requires at least 14 attributes with hidden layer width range as [1,31]. A much more suitable sample dimension will be 21 attributes with hidden layer width range as [1,11]. In the case of thyroid dataset 22 attributes is needed but it corresponds to a high width range [1,373]. This is computationally expensive and hence more suitable structure of thyroid dataset will be attribute size 45 against the width range [1,68]. 15 attributes is the minimum requirement for wine vintage dataset with hidden layer width range as [1,26]. A more appropriate one with lesser training complexity will be 20 against the range [1,10]. Hence more attributes needed to be identified depending on the sample size for reducing training complexity. Since increasing the attribute size is beyond the scope of this paper, it is sufficient to find the hidden layer width range corresponding to the dataset dimension given in table1. Table 3 gives the network width range on three given datasets. Experiment analysis done with Mathlab 2018(b) software to analyze whether the width of the hidden layer of the network that shows best performance falls within the range specified in table 3. This is experimentally studied on the basis of the performance plot obtained on each training. Bars with low MSE and Low Dissimilarity measure are considered to be optimal plot. The results obtained in table 4 are analysed with bar graphs shown in fig 2, fig 3, fig 4 and optimal hidden layer width is recorded in table 6.

**TABLE 5**
**ATTRIBUTE SIZE AND RESPECTIVE RANGE ON DIMENSION OF HIDDEN LAYER**

| Types of Glass Dataset (r=214) | | Thyroid Dataset (r=7200) | | Wine Vintage Dataset (r=178) | |
|---|---|---|---|---|---|
| Attribute size (n) | $[l_m, L_m]$ | Attribute size (n) | $[l_m, L_m]$ | Attribute size (n) | $[l_m, L_m]$ |
| 14 | [1,31] | 22 | [1,373] | 15 | [1,26] |
| 15 | [1,31] | 23 | [1,332] | 16 | [1,21] |
| 16 | [1,25] | 24 | [1,299] | 17 | [1,17] |
| 17 | [1,21] | 25 | [1,270] | 18 | [1,14] |
| 18 | [1,18] | 26 | [1,245] | 19 | [1,12] |
| 19 | [1,15] | 27 | [1,223] | 20 | [1,10] |
| 20 | [1,13] | 28 | [1,204] | 21 | [1,9] |
| 21 | [1,11] | 29 | [1,188] | 22 | [1,8] |
| 22 | [1,10] | 30 | [1,173] | 23 | [1,7] |
| 23 | [1,8] | 31 | [1,160] | 24 | [1,6] |
| 24 | [1,7] | 32 | [1,149] | 25 | [1,5] |
| 25 | [1,7] | 33 | [1,138] | 26 | [1,5] |
| 26 | [1,6] | 34 | [1,129] | 27 | [1,4] |
| 27 | [1,5] | 35 | [1,120] | 28 | [1,4] |
| 28 | [1,5] | 36 | [1,113] | 29 | [1,3] |
| 29 | [1,4] | 37 | [1,106] | 30 | [1,3] |
| 30 | [1,4] | 38 | [1,100] | 31 | [1,2] |
| 31 | [1,3] | 39 | [1,94] | 32 | [1,2] |
| 32 | [1,3] | 40 | [1,89] | 33 | [1,2] |
| 33 | [1,3] | 41 | [1,84] | 34 | [1,2] |
| 34 | [1,2] | 42 | [1,79] | 35 | [1,2] |
| 35 | [1,2] | 43 | [1,75] | 36 | 1 |
| 36 | [1,2] | 44 | [1,71] | 37 | 1 |
| 37 | [1,2] | 45 | [1,68] | 38 | 1 |
| 38 | [1,2] | 46 | [1,64] | 39 | 1 |
| 39 | 1 | 47 | [1,61] | 40 | 1 |
| 40 | 1 | 48 | [1,59] | 41 | 1 |
| 41 | 1 | 49 | [1,56] | 42 | 0 |
| 42 | 1 | 50 | [1,53] | 43 | 0 |

**TABLE 6**
**ANALYSIS ON THE EXPERIMENTAL STUDY**

| Dataset | Input Size | Sample Size | Optimal Network Width |
|---|---|---|---|
| Types of Glass | 9 | 214 | 1 |
| Thyroid | 21 | 7200 | 1,3,6 |
| Wine Vintage | 13 | 178 | 7 |

## 5. CONCLUSION

VC Dimension is the most widely used and acceptable measure that can quantify the complexity of functions in recognizing patterns computable by neural network. Since this concept is probabilistic in nature it lacks high intellectual decisions. If this concept is approached in an analytical way it can be applied theoretically to study the complexity of functions and structure the network accordingly. The paper analyses the VC Dimension in an $\varepsilon$−identifiable function space thereby creating a way to size the network according to sample dimension.

Theorems formulated in this paper provides conditions on the size of the sample space to have a minimum degree of training complexity with sufficient number of parameters. This eliminates the unnecessary computations due to large number of units in the hidden layer by limiting the hidden layer size within a finite bounds.

This could be further extended by the normalization of inputs and regularization of weights which provides a fine tuning to the optimization process. This technique can be generalized to any binary classification problem irrespective of sample distribution and size. The work is limited in its network architecture, binary classification problem and activation functions. Also the experimental analysis is limited to three datasets and more datasets with varying dimensions need to be experimented in different software techniques which is beyond the scope of this paper.